\documentclass[a4paper]{article}
\usepackage{iwslt18,amssymb,amsmath,epsfig}
\usepackage{times}
\usepackage{latexsym}
\usepackage{todonotes}
\usepackage{mathtools}
\usepackage{amsmath,bm}
\usepackage{multicol}
\usepackage{multirow}

\usepackage{mwe}
\usepackage{subfig}

\usepackage{xcolor,colortbl}
\definecolor{Gray}{gray}{0.85}

\usepackage{url}
\def\reg{{\rm\ooalign{\hfil
     \raise.07ex\hbox{\scriptsize R}\hfil\crcr\mathhexbox20D}}}

\title{Multitask Learning For Different Subword Segmentations In Neural Machine Translation}

 \makeatletter
 \def\name#1{\gdef\@name{#1\\}}
 \makeatother
 \name{{\em Tejas Srinivasan, Ramon Sanabria, Florian Metze}}

\address{Language Technologies Institute  \\
Carnegie Mellon University, USA \\
{\small \tt tsriniva, ramons, fmetze@cs.cmu.edu}
}
\begin{document}
\maketitle

\begin{abstract}

In Neural Machine Translation (NMT) the usage of subwords and characters as source and target units offers a simple and flexible solution for translation of rare and unseen words. 
However, selecting the optimal subword segmentation involves a trade-off between expressiveness and flexibility, and is language and dataset-dependent.
We present Block Multitask Learning (BMTL), a novel NMT architecture that predicts multiple targets of different granularities simultaneously, removing the need to search for the optimal segmentation strategy.
Our multi-task model exhibits improvements of up to 1.7 BLEU points on each decoder over single-task baseline models with the same number of parameters on datasets from two language pairs of IWSLT15 and one from IWSLT19.
The multiple hypotheses generated at different granularities can be combined as a post-processing step to give better translations, which improves over hypothesis combination from baseline models while using substantially fewer parameters.

\end{abstract}

\begin{figure}[t]
\centering
\includegraphics[width=0.4\textwidth]{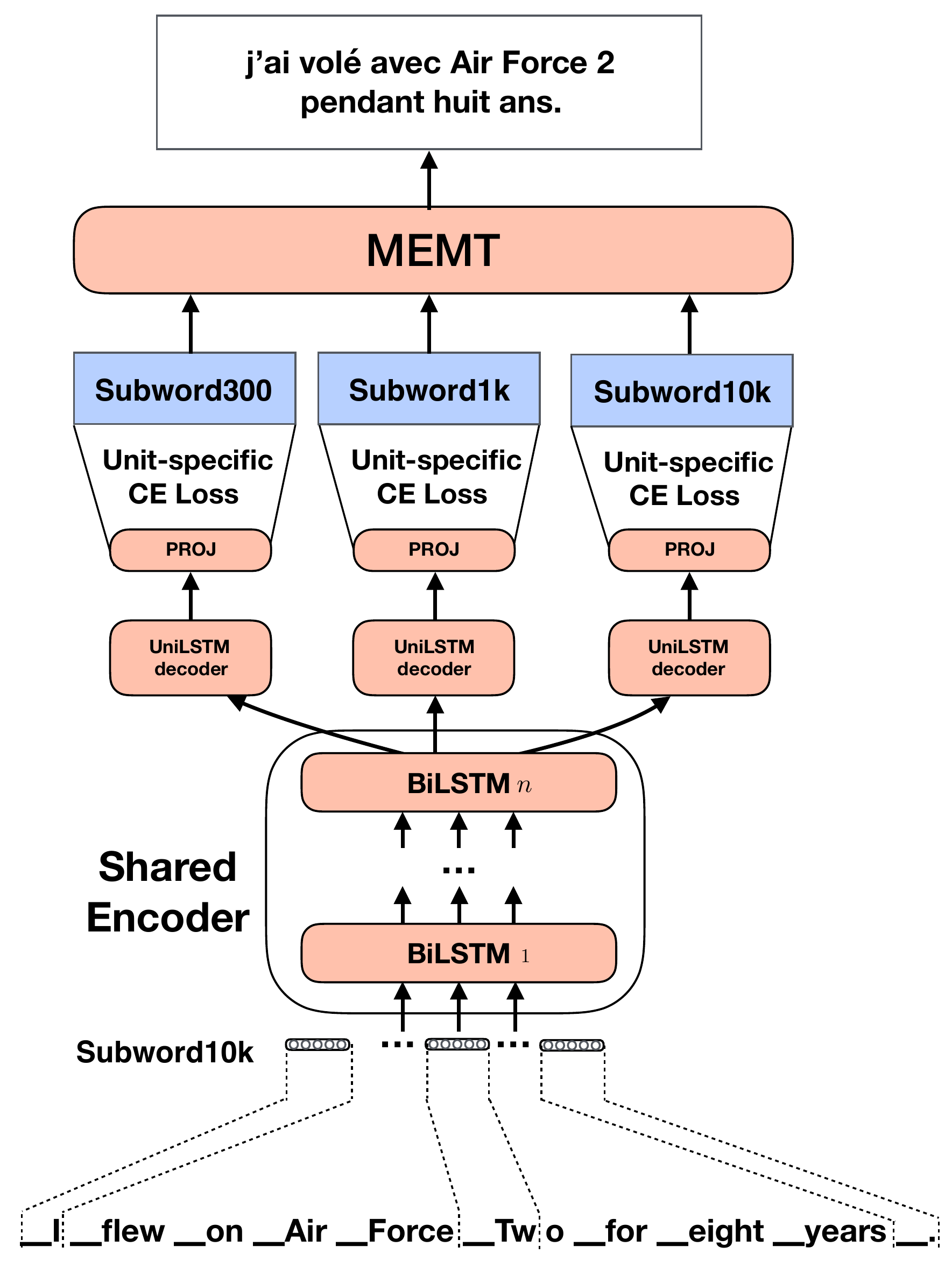}
 \caption{Our Block Multitask Learning (BMTL) Model learns to translate the same sentence in different subword-level units with multiple granularities at the same time. Finally, a Multi-Engine Machine Translation (MEMT) system combines all of them.}\label{label:bmtl}
 \end{figure}

\section{Introduction}
\label{sec:intro}



Neural Machine Translation (NMT) \cite{forcada1997recursive, cho2014learning, sutskever2014sequence} provides a simple, end-to-end framework for translating text from one language to another. NMT approaches have largely outperformed and replaced previous statistical translation methods. Traditionally, NMT systems used words as source and target units, which have three main disadvantages. First, word-based models are unable to translate rare and out of vocabulary (OOV) words in the source language. Second, they can not produce unseen target words, such as morphological variants of observed words (\textit{e.g.}, deriving realistic from real). Third, they have to handle large source and target language vocabularies (\textit{i.e.}, large look-up matrices), which makes them less scalable in term of computation and memory. A large vocabulary also implies data sparsity where the number of tokens is not balanced.


A common solution for the problems mentioned above is to perform word segmentation. The Byte-Pair Encoding (BPE) algorithm \cite{sennrich2016neural} groups units together according to their frequency. By presetting the desired vocabulary size, the BPE algorithm generates a segmentation of the data by representing words as a collection of subword units. More recently, \cite{kudo2018subword} proposed a forward probabilistic subword segmentation algorithm which is based on a unigram language model, in contrast to the deterministic BPE algorithm. Unlike BPE, the unigram language model is capable of providing multiple subwords with probabilities, thereby making the segmentation more flexible. However, unigrams did not exhibit significant improvement over BPE. 

Subword-level NMT systems resolve the drawbacks of word-level models by providing open-vocabulary capabilities to the model and reducing the vocabulary size considerably. However, subword-level systems also have some drawbacks. Most importantly, input and target sequences are longer, which makes them slower to train and decode, and implies long-range dependencies which are difficult to model \cite{chapter-gradient-flow-2001}. Also, the open-vocabulary capabilities of the model might generate undesired variants of correct translations. Finally, subword embeddings do not carry as much semantic information as words and therefore modeling this information becomes much more difficult.

All units discussed above present a trade-off between flexibility and semantic information (\textit{i.e.}, characters are more flexible with less semantic information, whereas words can not translate OOV words but contain more semantic information). This trade-off makes the selection of optimal segmentation a non-trivial problem, for a given dataset and language pair. Generally, the optimal segmentation is treated as a hyper-parameter that needs to be found by brute-force search, and this search is time-consuming and error-prone. This problem is even more emphasized in multilingual settings where the optimal segmentation needs to be found for each language \cite{johnson2017google}.

In this paper, we propose a block multitask learning (BMTL) model that, by using multiple subword segementations in the target domain, translates the same input with different granularities (see Fig. \ref{label:bmtl}). All hypotheses are combined posteriorly with Multi-Engine Machine Translation (MEMT) system~\cite{DBLP:conf/wmt/HeafieldL11}, which generates the final hypothesis of the system. Our experiments show that, in general, each output segmentation of our BMTL outperforms all single task approaches that use the same number of parameters. By combining the outputs of BMTL with MEMT, our system still outperforms the combination of single-task models, in spite of using lesser parameters in total. We hypothesize that sharing an encoder among different decoder-specific granularities, makes the encoded representation more general and robust, which yields a better translation and therefore an improvement in BLEU score.


The main contributions of this paper are as follows:
\begin{itemize}
    \item We introduce Block Multitask Learning (BMTL), an NMT framework that, by using multiple subword segmentations in the target domain, translates the same input with different granularities (see Fig. \ref{label:bmtl}) (Section ~\ref{subsec:bmtl}).
    \item We present a set of experiments in three different IWSLT language pairs (En-\{Fr, Vi, Cs\}) that show improvements on each output segmentation of BMTL, outperforming all single task approaches that use the same number of parameters. (Section~\ref{sec:results})
    \item We show that by combining the outputs of BMTL with Multi-Engine Machine Translation (MEMT)~\cite{DBLP:conf/wmt/HeafieldL11}, our system still outperforms the combination of single-task models, in spite of using fewer parameters in total (Section \ref{sec:memt-results}).
\end{itemize}




\section{Architecture}
\label{sec:bmtl}


In this section, we will first introduce our baseline model that consists of a standard attention-based encoder-decoder model~\cite{Bahdanau2014} (Section~\ref{sec:baseline}). After that, we present our new BMTL architecture that uses the encoder-decoder model as the main building block (Section~\ref{subsec:bmtl}). Finally, we describe MEMT, the mechanism that we use for combining multiple hypotheses.

\subsection{Baseline Model}

\label{sec:baseline}

Our baseline model is a standard encoder-decoder model with a multilayer perceptron attention and tanh activation~\cite{Bahdanau2014}. The encoder is a bidirectional recurrent neural network with Gated Recurrent Units (BiGRU). This block of the system encodes the input subword embeddings. The decoder is also a recurrent neural network, but it uses Conditional GRU decoder~\cite{nematus}. The decoder, conditioned on the previously generated state and each encoded vector, generates an attention matrix that weighs all the hidden states generated by the encoder. The decoder continuously generates symbols until the end-of-sentence symbol is produced. This model can use different subword segmentations in the source and target space. We will refer to this henceforth as the baseline model.

\subsection{Block Multitask Learning}
\label{subsec:bmtl}

In BMTL (see Fig.~\ref{label:bmtl}), we extend the baseline model with a multitask learning approach. More specifically, in this case, each task is the generation of the translation in the target language, in different granularities. All tasks share the same encoder as in the baseline model. The encoded matrix is processed by multiple decoders, all of which have the same architecture as the baseline decoder. Each of the decoders has its own attention and set of parameters.

More formally, a BMTL model with decoders outputting units of BPE300, BPE1000 and BPE10000, can be written as

\begin{equation}
\begin{split}
\bm{e}_0 &= \text{Shared\_Encoder}(\bm{X})\\
\bm{S}_{\text{bpe300}} &= \text{CGRU\_bpe300}(\bm{e}_0)\\
\bm{S}_{\text{bpe1k}} &= \text{CGRU\_bpe1k}(\bm{e}_0)\\
\bm{S}_{\text{bpe10k}}&=  \text{CGRU\_bpe10k}(\bm{e}_0)\\
\end{split}
\end{equation}

where $\bm{S_n}$ is the generated hypothesis and $\text{CGRU\_n}$ is a decoder for the subword segmentation $n$.

During training, the losses obtained by all the decoders are normalized according to length, summed and averaged. We found that this approach works better than backpropagating each loss independently through its own decoder as well as the shared encoder. This allows the encoder to learn more generalized representations which are independent of the output subword granularity.

It is important to note that BMTL is a model agnostic technique and it can be easily ported to other architectures such as Transformer \cite{vaswani2017attention}.

\begin{table*}[t]
    \centering
  \begin{tabular}{| c | c | c c c | c c c |}
    \hline
    \multirow{2}{*}{Corpus} & \multirow{2}{*}{Model} & \multicolumn{3}{c|}{BMTL1} & \multicolumn{3}{c|}{BMTL2} \\ \cline{3-8}
    & & BPE300 & BPE1K & BPE10K & BPE10K & BPE16K & BPE32K \\ 
    \hline \hline 
     \multirow{2}{*}{En-Fr} & \cellcolor{gray!20} Baseline & \cellcolor{gray!20}35.6 & \cellcolor{gray!20}35.1 & \cellcolor{gray!20}\textbf{36} & \cellcolor{gray!20}35 & \cellcolor{gray!20}36 & \cellcolor{gray!20}34.8 \\
     & BMTL & \textbf{36} & \textbf{35.6}& 35.7 & \textbf{36.5} & \textbf{36.1} &  \textbf{36.5} \\ \hline  \hline
     \multirow{2}{*}{En-Vi} &  \cellcolor{gray!20}Baseline & \cellcolor{gray!20}26.4 & \cellcolor{gray!20}27.1 & \cellcolor{gray!20}26.3 & \cellcolor{gray!20}27.6 & \cellcolor{gray!20}27 & \cellcolor{gray!20}27.3\\ 
     & BMTL & \textbf{27} & \textbf{27.7}& \textbf{27.6} & \textbf{27.8} & \textbf{27.5} & \textbf{27.6} \\ 
     \hline  \hline
     \multirow{2}{*}{En-Cs}  & \cellcolor{gray!20} Baseline & \cellcolor{gray!20}17 & \cellcolor{gray!20}16.5 & \cellcolor{gray!20}16.7 & \cellcolor{gray!20}\textbf{16.7} & \cellcolor{gray!20}16.4 & \cellcolor{gray!20}\textbf{16.4} \\
     & BMTL & \textbf{17.6} & \textbf{17.7}& \textbf{17.4} & 16.6 & \textbf{16.8} & 16.3 \\ \hline
  \end{tabular}

  \caption{BLEU scores of our BMTL1 (\textit{i.e.}, BMTL combining BPE 300, BPE1K and BPE10K) and BMTL2 (\textit{i.e.}, BMTL combining BPE 10K, BPE16K and BPE32K) models, as well as the baseline models, on each of our three IWSLT language pairs (\textit{i.e.}, English to \{French, Czech, Vietnamese\}).}
  \label{tab:main-results}
\end{table*}

\subsection{Multi-Engine Machine Translation}
\label{subsec:memt}

As a post-processing step, our system uses MEMT to combine all generated hypotheses~\cite{DBLP:conf/wmt/HeafieldL11}. MEMT uses a variant of the METEOR aligner to align all the results of each decoder. It applies four constraints to generate the combined hypothesis. First, the sentences must start with start-of-sentence symbol and end-of-sentence symbol. Second, a token is used only once. Third, it forces weak monotonicity between the alignments, preventing too many jumps from the search algorithm. Fourth, it forces the completeness of the combined hypothesis by not skipping tokens unless the sentence ends. It is important to note that MEMT uses tokenized-word hypotheses as input.

\section{Experiments}
\label{sec:exp}

\subsection{Datasets and Preprocessing}
\label{sec:datapreproc}

We report results on two language pairs from the IWSLT 2015 TED Talks corpus - English to \{French, Vietnamese\}. We use the \texttt{tst2012} and \texttt{tst2013} sets as our development and testing sets. Furthermore, we also report results on the IWSLT 2019 text translation task from English to Czech. We use the provided training and development sets, and the \texttt{tst-COMMON} set for testing.

For preprocessing, each corpus is normalized, tokenized and truecased using Moses \cite{koehn2007moses}. Each corpus is then segmented to different BPE vocabulary sizes using the sentencepiece implementation\footnote{\url{https://github.com/google/sentencepiece}}.


\subsection{Implementation Details}
\label{sec:impexp}
All models are trained using Adam optimizer \cite{DBLP:journals/corr/KingmaB14}, with a learning rate of 0.0001, decay of 0.9 and batch size of 32. All models have 2 layers of bidirectional encoders of size 512 in each direction, decoder of size 1024, and input and output embeddings of size 512. We also apply dropout with probability 0.1 in the encoder and decoder. The norm of the gradient is clipped with a threshold of 1 \cite{DBLP:journals/corr/abs-1211-5063}. All models are implemented using the nmtpytorch framework ~\cite{nmtpy2017}.
The output hypotheses are detokenized and detruecased using Moses, before using sacreBLEU \cite{post2018call}
for scoring the translations. Finally, all hypothesis are combined with the MEMT implementation\footnote{\url{https://github.com/kpu/MEMT}} 
 with the default configuration provided.

\subsection{Results}
\label{sec:results}

We experiment with two variants of the BMTL model. BMTL1 has inputs of BPE10K and decoders of BPE300, BPE1K and BPE10K (as seen in Figure \ref{label:bmtl}). BMTL2 has inputs of BPE32K and decoders of BPE10K, BPE16K and BPE32K. We experiment with different input segmentations to show that our architecture shows improvements irrespective of the input unit. For each BMTL model, we also train three baseline encoder-decoder models - each with the same input units as BMTL and an output corresponding to one of the BMTL decoders. For instance, we compare the output of BMTL1's BPE300 decoder with an encoder-decoder model that has input units of BPE10K and output units of BPE300.

Table \ref{tab:main-results} shows the results of our experiments on the BMTL1 and BMTL2 models, as well as the baseline models. We observe that almost all of our BMTL decoders (in both BMTL1 and BMTL2) outperform the corresponding baseline models across all three languages, with an improvement of upto 2 BLEU points. This exhibits our architecture's ability to learn more robust encoded representations, irrespective of language, input units, and combination of output segmentations. \footnote{All of our baseline numbers are comparable to numbers on the same datasets in \cite{wolk2015pjait}.}

\begin{table}[ht]
  \centering
  \scalebox{1}{
  \begin{tabular}{| c | c | c | c|}
    \hline
    Corpus & Model & BLEU &  \# param. (M)  \\
    \hline
    \multirow{4}{*}{} En-Fr  &  \cellcolor{gray!20}Baseline1 & \cellcolor{gray!20}37.4 & \cellcolor{gray!20}92.03 \\
         & BMTL1 & \textbf{37.5}  & 71.82 \\ \cline{2-3} 
       & \cellcolor{gray!20}Baseline2 & \cellcolor{gray!20}37.3 & \cellcolor{gray!20}149.78 \\
          & BMTL2 & \textbf{37.8} &  114.85  \\ \hline
         \multirow{4}{*}{} En-Vi  &  \cellcolor{gray!20}Baseline1 & \cellcolor{gray!20}28.8 & \cellcolor{gray!20}92.03 \\
         & BMTL1 & \textbf{29} &  71.82 \\ \cline{2-3}
       & \cellcolor{gray!20}Baseline2 & \cellcolor{gray!20}28.5 & \cellcolor{gray!20}149.78 \\
          & BMTL2 & \textbf{29} &  114.85 \\ \hline
        \multirow{4}{*}{} En-Cs  &  \cellcolor{gray!20} Baseline1 & \cellcolor{gray!20}18.1 & \cellcolor{gray!20}92.03 \\
         & BMTL1 & \textbf{18.7} &  71.82  \\ \cline{2-3}
        & \cellcolor{gray!20}Baseline2 & \cellcolor{gray!20}\textbf{18.1} & \cellcolor{gray!20}149.78  \\
          & BMTL2 & 18 &  114.85  \\ \hline
  \end{tabular}
  }
  \caption{BLEU scores using MEMT. We compare two systems. First, BMTL1 with Baseline1, that combines 300, 1K and 10K systems in BMTL and independently-trained baselines respectively. Second, BMTL2 with Baseline2 that combines 10k, 16k and 32K systems in BMTL and baselines respectively. \# param. lists the number of trainable parameters in (M)illions. For baseline MEMT, we report the sum of the parameters of the independently-trained baselines.}
  \label{tab:memt}
\end{table}

\begin{figure*}[t]%
 \centering
 \subfloat[BLEU scores from BPE300 decoder]{\includegraphics[width=.45\linewidth]{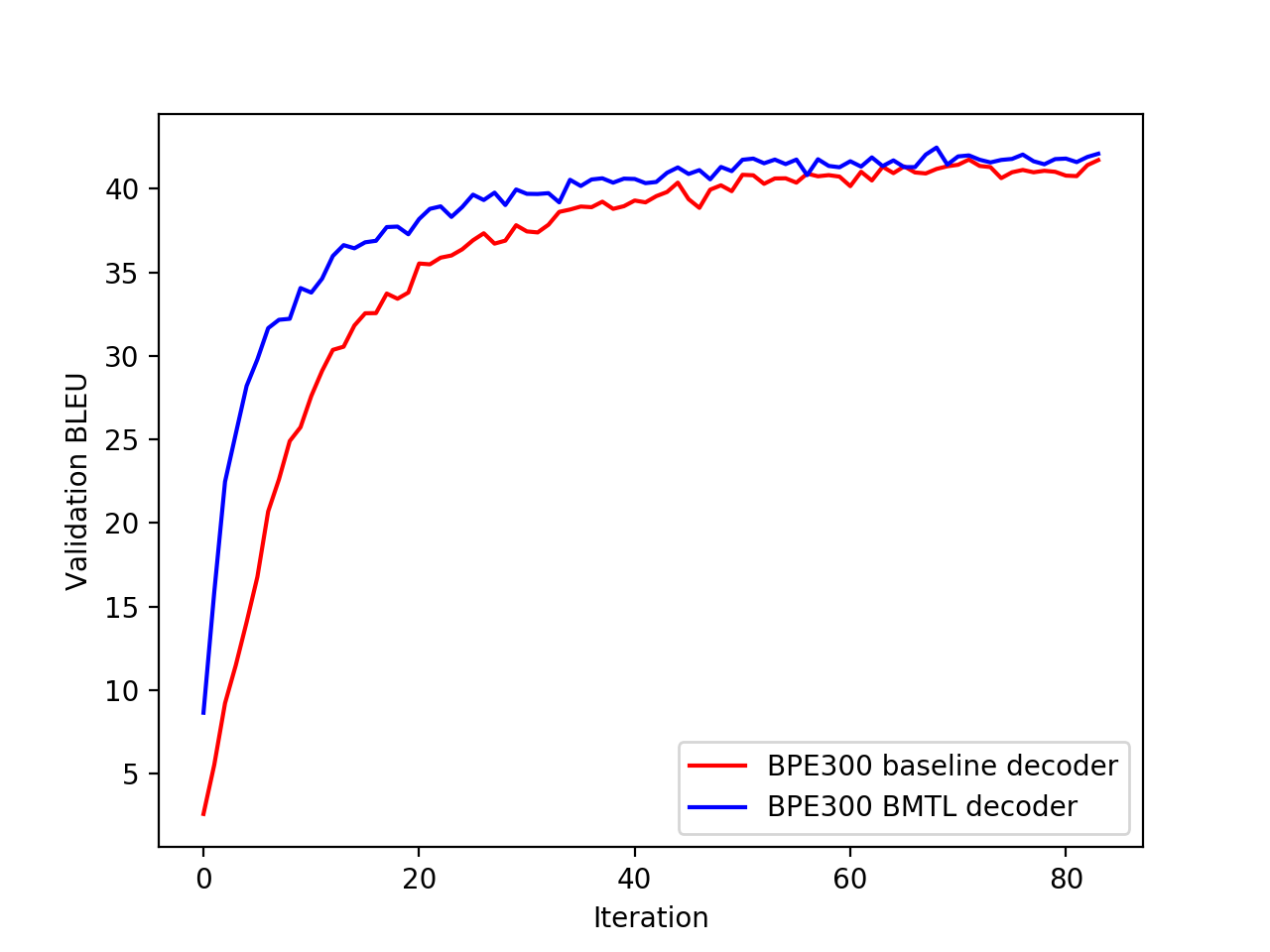}\label{fig:bpe300}}%
 \subfloat[BLEU scores from BPE1K decoder]{\includegraphics[width=.45\linewidth]{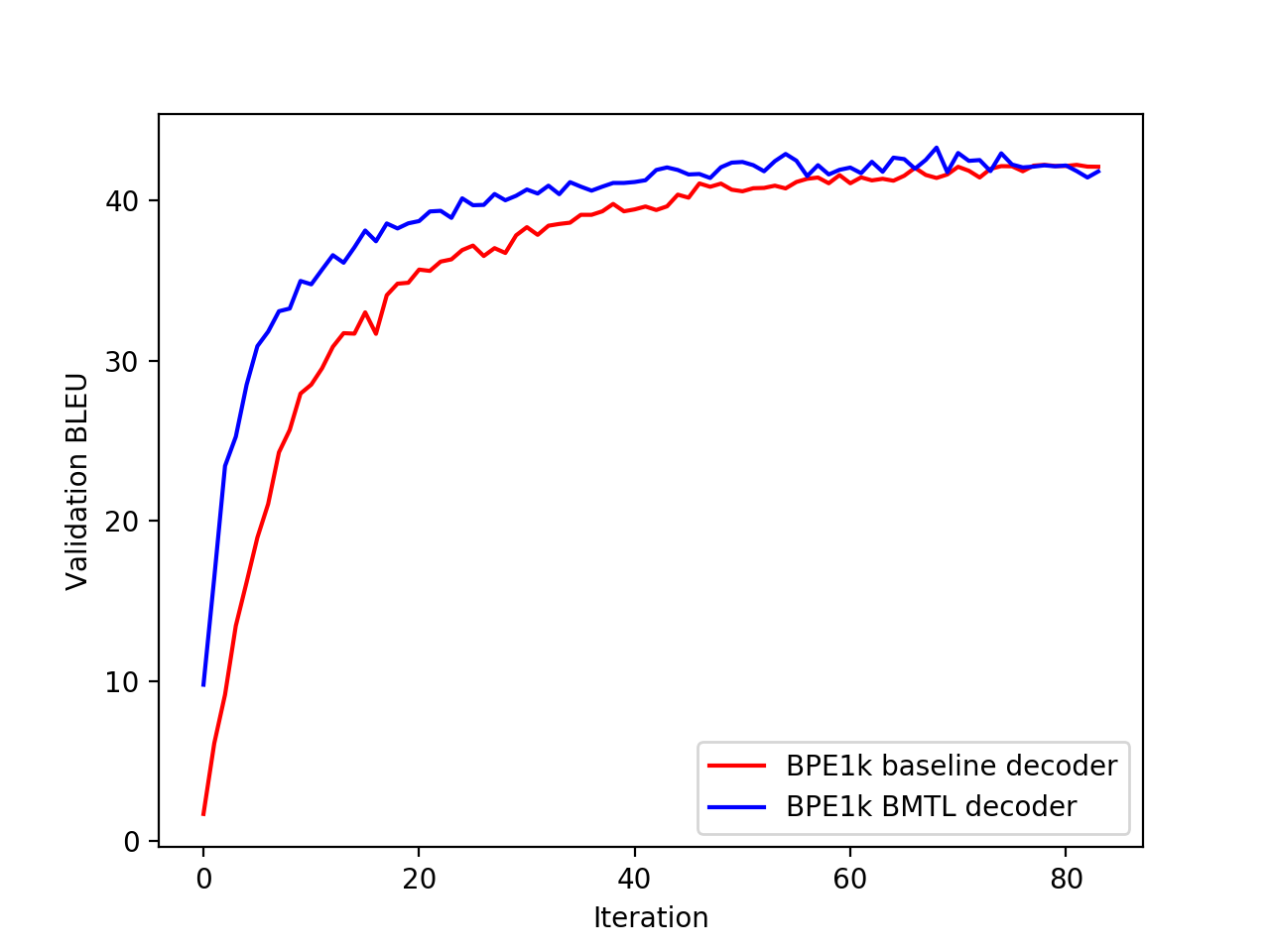}\label{fig:bpe1k}}\\
 \subfloat[BLEU scores from BPE10K decoder]{\includegraphics[width=.45\linewidth]{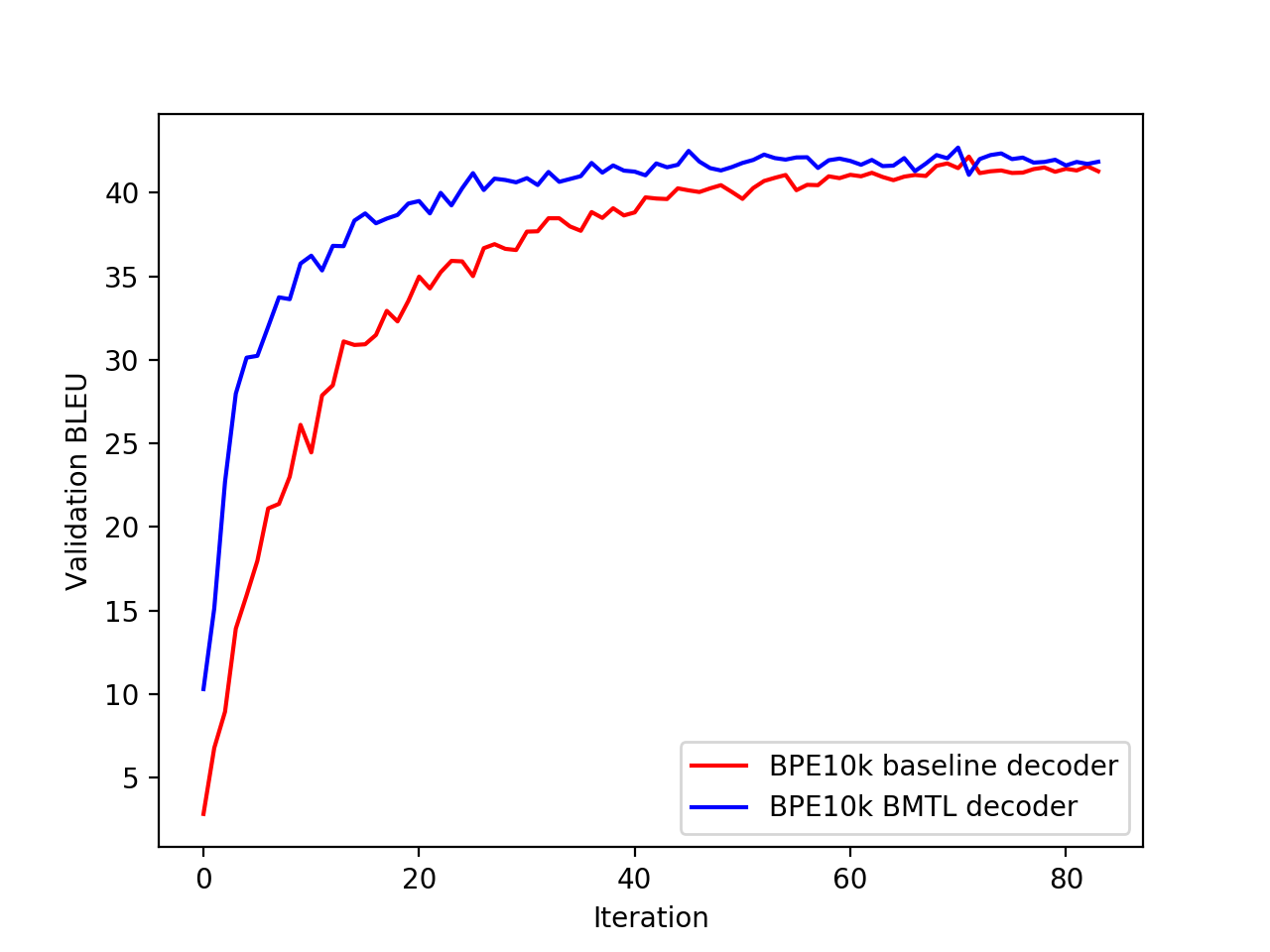}\label{fig:bpe10k}}%
 \caption{Number of iterations versus validation BLEU scores for decoders in our BMTL1 and Baseline1 models, on the IWSLT15 En-Fr dataset. We compare the number of iterations to converge between each of the decoders in our BMTL1 model and the corresponding Baseline1 model}%
 \label{fig:bleu-var}%
\end{figure*}

These improvements are on models that have the same size as the baselines. Although at training time, the model includes multiple decoders and a shared encoder, while testing, we need to utilize only a single decoder and encoder, thus making it comparable to the baseline models. Each of our BMTL decoders also converges faster than the corresponding individual baseline models (Figure \ref{fig:bleu-var}).

\subsection{Hypothesis Combination}
\label{sec:memt-results}

We explore the possibility of combining hypotheses from each of the decoders in BMTL (see Fig.~\ref{label:bmtl}). We use Multi-Engine Machine Translation (MEMT) \cite{DBLP:conf/wmt/HeafieldL11}, to get a single hypothesis by combining the hypotheses of each BMTL decoder, the results of which can be seen in Table \ref{tab:memt}.

We see that the MEMT combination from the BMTL models almost always outperforms the baselines. Although these gains are slightly lower than the ones achieved by the individual BMTL decoders, they are achieved using models with significantly fewer parameters (greater than 20\% reduction from the combined baseline models). This reduction is because the combined baseline models have multiple encoders, whereas BMTL has a shared encoder for each of the decoders.

\section{Related Work}


Multiple previous works have analyzed the differences between target units of different subword resolution in NMT systems. These works make the observation that the optimal segmentation depends on three elements: number of OOV words~\cite{sennrich2016neural}, language~\cite{kudo2018subword}, and size of the model~\cite{cherry2018revisiting}. Each of these dependencies makes the task of finding the optimal subword segmentation computationally infeasible and prone to error.



In an attempt to solve this problem, \cite{kreutzer2018learning} propose a dynamic end-to-end, data-driven segmentation. \cite{kreutzer2018learning} uses the Adaptive Computation Time paradigm~\cite{graves2016adaptive} to let the network learn an optimal segmentation. This approach, however, does not match nor overcome the BLEU score of any of the manual segmentations proposed. Our work, instead, benefits from having mutliple representations for the same input outperforming almost all single-segmentation baselines proposed.

To take advantage of the different available segmentations and solve the problem of OOV words, \cite{luong2016achieving} proposes a hybrid system that combines words and character-based models. Translation occurs primarily at the word level, and the system uses the character-level model when an unknown symbol is predicted. \cite{luong2016achieving} is similar to our approach in its usage of multiple target unit segmentations. Even though ~\cite{luong2016achieving} achieves similar improvements to ours, the incorporation of a second character-based architecture makes their approach more memory intensive and slow than the baseline model (\textit{i.e.}, word-based NMT) at test time.

Perhaps more related to our work, there are two recent NMT approaches that combine multiple BPE segmentations. First, \cite{salesky2018optimizing} sequentially increase the number of units during training each time the architecture converges. This method achieves comparable results to grid search, without the need of training the model a number of times. Second, \cite{makoto2018improving} proposes summing the multiple subword embeddings from different segmentations to the same embedding layer. Although \cite{salesky2018optimizing,makoto2018improving} propose to use multiple target segmentations in the same system, there are multiple differences from our work. Even though \cite{salesky2018optimizing} finds the optimal segmentation, they do not use concurrently different target units. ~\cite{makoto2018improving} uses multiple representations in parallel in the input while we use ours in the output.

Related to the architecture of our model, ~\cite{caruana1997multitask} proposed the first work on multitask learning. Similarly, many approaches integrated other tasks in NMT models such as the translation of more languages~\cite{daxiang2015multi}, Part-Of-Speech tagging~\cite{niehues2017exploiting} or general syntax~\cite{kiperwasser2018scheduled}. However, none of them combined different granularities of the same sentence. More recently, in Automatic Speech Recognition, \cite{sanabria2018hierarchical} proposed also the use of multiple levels of segmentation in a hierarchical multitask learning structure. The improvements showed in \cite{sanabria2018hierarchical} inspired this paper.

\section{Conclusions And Future Work}
We propose Block Multitask Learning (BMTL) model that translates the same input to multiple subword granularities in the target language and is trained in a multi-task learning fashion. Our BMTL decoders outperform the single-task baseline models across all languages, for different input units, and different combinations of output segmentations, while having comparable model size. We also use Multi-Engine Machine Translation to combine multiple decoders' hypotheses as a post-processing step; this also achieves improvements over combining single-task hypotheses, despite having significant fewer parameters. 


With regards to how we can expand this work in the future, we are investigating several potential future directions for BMTL. First, we are investigating how we can effectively encode source sentences using different input segmentations (similar to \cite{passban2018improving}). Second, we are exploring if weighing each loss function provided by each decoder can help the model to learn a better representation. Third, we are exploring an online beam-search strategy that uses the hypothesis of all decoders. This technique will constitute a more elegant solution for joint decoding than MEMT. On a similar note, we want to explore an online decoding strategy during training that can selectively switch between decoders. Finally, while our models have so far been tested on small-scale IWSLT datasets, we plan to investigate the performance of our model on larger WMT datasets, and using the state-of-the-art Transformer network.
\bibliographystyle{IEEEtran}
\bibliography{acl2018}
\bibliographystyle{acl_natbib}

\end{document}